\documentclass[sigconf]{acmart}
\AtBeginDocument{%
  }

\usepackage{algorithm}
\usepackage{algpseudocode}
\usepackage{amsmath}
\usepackage{multirow}
\usepackage{graphicx}
\usepackage{colortbl}

\usepackage{pifont}  
\begin{document}

\title{Enhancing the Geometric Problem-Solving Ability of Multimodal LLMs via Symbolic-Neural Integration}



\author{Yicheng Pan}
\affiliation{%
  \institution{University of Science and Technology of China}
  \city{}
  \country{}
}
\email{yichpan@mail.ustc.edu.cn}

\author{Zhenrong Zhang}
\affiliation{%
  \institution{University of Science and Technology of China}
  \city{}
  \country{}
}
\email{zzr666@mail.ustc.edu.cn}

\author{Pengfei Hu}
\affiliation{%
  \institution{University of Science and Technology of China}
  \city{}
  \country{}
}
\email{pengfeihu@mail.ustc.edu.cn}

\author{Jiefeng Ma}
\affiliation{%
  \institution{University of Science and Technology of China}
  \city{}
  \country{}
}
\email{jfma@mail.ustc.edu.cn}

\author{Jun Du}
\authornote{Corresponding author.}
\affiliation{%
  \institution{University of Science and Technology of China}
  \city{}
  \country{}
}
\email{jundu@ustc.edu.cn}

\author{Jianshu Zhang}
\affiliation{%
  \institution{iFLYTEK Research}
  \city{}
  \country{}
}
\email{jszhang6@iflytek.com}

\author{Quan Liu}
\affiliation{%
  \institution{iFLYTEK Research}
  \city{}
  \country{}
}
\email{quanliu@iflytek.com}

\author{Jianqing Gao}
\affiliation{%
  \institution{iFLYTEK Research}
  \city{}
  \country{}
}
\email{jqgao@iflytek.com}

\author{Feng Ma}
\affiliation{%
  \institution{iFLYTEK Research}
  \city{}
  \country{}
}
\email{fengma@iflytek.com}


\renewcommand\footnotetextcopyrightpermission[1]{}
\settopmatter{printacmref=false}
\setcopyright{none}

\begin{abstract}
  Recent advances in Multimodal Large Language Models (MLLMs) have achieved remarkable progress in general domains and demonstrated promise in multimodal mathematical reasoning. However, applying MLLMs to geometry problem solving (GPS) remains challenging due to lack of accurate step-by-step solution data and severe hallucinations during reasoning. In this paper, we propose GeoGen, a pipeline that can automatically generates step-wise reasoning paths for geometry diagrams. By leveraging the precise symbolic reasoning, \textbf{GeoGen} produces large-scale, high-quality question-answer pairs. To further enhance the logical reasoning ability of MLLMs, we train \textbf{GeoLogic}, a Large Language Model (LLM) using synthetic data generated by GeoGen. Serving as a bridge between natural language and symbolic systems, GeoLogic enables symbolic tools to help verifying MLLM outputs, making the reasoning process more rigorous and alleviating hallucinations. Experimental results show that our approach consistently improves the performance of MLLMs, achieving remarkable results on benchmarks for geometric reasoning tasks. This improvement stems from our integration of the strengths of LLMs and symbolic systems, which enables a more reliable and interpretable approach for the GPS task. Codes are available at \url{https://github.com/ycpNotFound/GeoGen}.

\end{abstract}

%
%

\ccsdesc[500]{Computing methodologies~Reasoning about belief and knowledge}
\ccsdesc[300]{Computing methodologies~Theorem proving algorithms}
\ccsdesc[100]{Computing methodologies~Neural networks}


\keywords{Geometry problem solving, Multimodal large language models, Symbolic reasoning}
\begin{teaserfigure}
  \includegraphics[width=\textwidth]{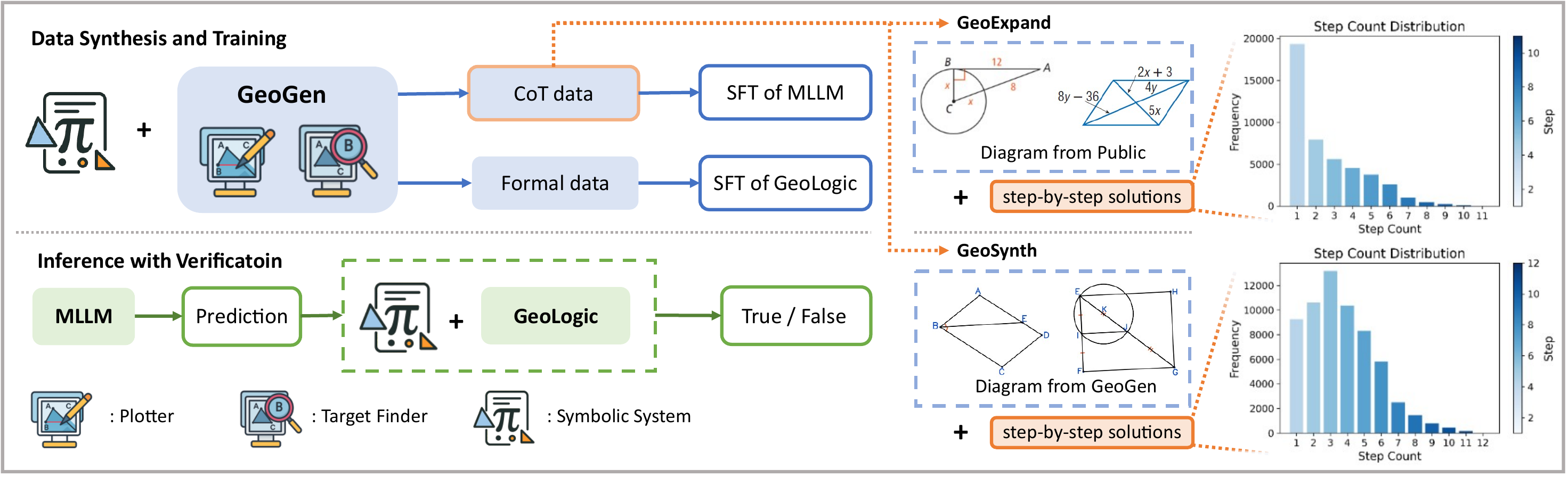}
  \caption{Framework of our proposed integration methods of symbolic system and MLLMs.}
  \Description{Enjoying the baseball game from the third-base
  seats. Ichiro Suzuki preparing to bat.}
  \label{fig:teaser}
\end{teaserfigure}


%
\settopmatter{printfolios=true}
\maketitle

\section{Introduction}

Recent advances in Multimodal Large Language Models (MLLMs) \cite{bai2025qwen2,hurst2024gpt, chen2024internvl, liu2024llavanext} have demonstrated remarkable progress across a wide range of general domains. Leveraging their inherent reasoning capabilities, these models have also shown promising potential in multimodal mathematical reasoning \cite{peng2024multimath,shi2024mathllava}. However, their application to Geometry Problem Solving (GPS) remains challenging. As a long-standing problem in the AI community, GPS represents a complex form of multimodal mathematical reasoning \cite{lu2021inter, chen2021geoqa, zhang2023multi}. In a typical plane geometry problem, an automatic solver is required to produce the correct answer or reasoning process based on a given diagram and textual question. This task demands fine-grained visual understanding (e.g., identifying points, lines and angles) \cite{zhang2022plane} and multi-step theorem-based reasoning \cite{lu2021inter}, posing unique challenges distinct from other mathematical reasoning tasks.

Existing MLLM solvers for GPS primarily rely on supervised fine-tuning (SFT) paradigms using chain-of-thought (CoT) supervision \cite{shi2024mathllava,zhuang2024math, wei2022chain, liang2023unimath}. However, high-quality reasoning processes in natural language are scarce in public geometry datasets \cite{lu2021inter, peng2024multimath}, which poses a significant bottleneck to the effective SFT of MLLMs, and limits their reasoning capabilities. While existing approaches attempt to construct more CoT data through template-based generation \cite{kazemi2023geomverse, zhang2024mavis} or large language model (LLM) synthesis (e.g. GPT-4o) \cite{gao2023gllava, cai2024geogpt4v, hurst2024gpt, yang2024qwen2}, significant challenges persist. Template-driven methods suffer from rigid formatting constraints that restrict content diversity and real-world applicability \cite{deng2024rcot}. Meanwhile, solutions generated by LLMs exhibit inherent uncertainty in output quality.

On the other hand, symbolic-based approaches for GPS have demonstrated strong effectiveness \cite{lu2021inter,wu2024egps,peng2023geodrl}. Through specified geometry formal language, symbolic systems can conduct precise symbolic reasoning to derive solutions via searching and executing geometry theorems in a structured manner \cite{zhang2023formalgeo}. However, such methods heavily rely on manually crafted rules, limiting their scalability and adaptability \cite{deng2024rcot}. Some recent efforts have attempted to combine LLMs with symbolic systems, where the general knowledge of LLMs are used to guide the symbolic search process \cite{zhao2025pigps,AlphaGeometryTrinh2024,chervonyi2025gold}. Nonetheless, these approaches primarily enhance the symbolic system, rather than improving the intrinsic geometric problem-solving abilities of the language model itself. Overall, the integration of symbolic systems and LLMs remains an underexplored direction in the context of geometry problem solving.

In this work, we find that the in MLLMs with symbolic systems can effectively alleviate the data scarcity issue discussed above. To be specific, the symblic system are capable of applying search algorithms and uncover a wide range of potential geometric relationships within a single diagram \cite{zhang2023formalgeo}. This insight enables the expansion of a single geometry image into multiple problem-solving targets, thereby increasing the effective volume of training data. Furthermore, due to strong language modeling capabilities, LLMs can serve as a bridge between natural language and formal language \cite{pan2023logic}. This translation capability serves as a foundation for integrating the strengths of both symbolic reasoning and vision-language understanding. 

Based on the above observations, we propose \textbf{GeoGen}, a novel pipeline that automatically construct geometric multi-step CoT data leveraging both LLM and symbolic system. Our pipeline enables automatic synthesis of geometry diagrams based on formal language, as well as automatic annotation for both synthesized and public images with multi-step reasoning data. GeoGen first performs precise symbolic reasoning and generate reasoning paths with guaranteed correctness. Next, GeoGen leverages the external LLMs to translate structured reasoning paths into coherent natural language explanations. Based on this pipeline, we further annotate and synthesize two datasets: GeoExpand and GeoSynth. GeoExpand dataset is created with 45k solutions by expanding upon the images from existing datasets \cite{lu2021inter,zhang2023multi}. Additionally, we construct GeoSynth dataset from scratch producing 62k images along with Q\&A pairs to further increase the training scale and diversity. Compared to other geometry diagram synthesis methods \cite{zhang2024mavis,kazemi2023geomverse}, our constructed datasets incorporate richer geometric knowledge and more detailed multi-step reasoning, providing stronger supervision for training geometry-solving models.

To further enhance the logical reasoning capabilities of MLLMs in GPS, we view symbol systems as function tools for MLLMs, and introduce \textbf{GeoLogic}. Leveraging the general language capabilities of LLMs \cite{wu2024survey,ahn2024large,jiang2024survey,zhao2023survey}, GeoLogic is trained on formal–natural language solution pairs generated by GeoGen, and learns to translate natural language solutions into formal representations required by symbolic solvers. This capability equips the model to treat symbolic systems as external tools that can be invoked to support reasoning. In practice, we take an initial step toward this integration by using symbolic systems to verify MLLM predictions step by step. This verification process enforces alignment between the model’s outputs and geometric principles, making the reasoning more reliable. 

Our overall framework of integration methods is illustrated in Figure ~\ref{fig:teaser}. During the training phase, we conduct extensive SFT experiments using public and synthetic data. Comprehensive evaluations of SFT on our datasets validate their effectiveness, showing consistent improvements across various MLLMs and achieving remarkable performance across multiple geometry benchmarks. In the inference phase, we leverage GeoLogic along with the verification capabilities of the symbolic system. A tree-based search is performed, which enables step-by-step validation on MLLM's prediction, reducing reasoning errors and hallucinations.

\begin{enumerate}
  \item We propose GeoGen, a symbolic-based pipeline that can automatically generate step-by-step solutions for both existing datasets and synthetic diagrams.  
  \item We further construct GeoExpand and GeoSynth dataset, containing 45K and 62K questions respectively, each accompanied by multi-step reasoning solutions.
  \item We introduce GeoLogic, which enables the integration of symbolic systems as verifiers during MLLM inference, enhancing the rigor and reliability of the reasoning process.
  \item Experimental results demonstrate the effectiveness of our approach in both training and inference phases.
\end{enumerate}

\section{Related Work}
\subsection{Geometry Problem Solving}
Geometry Problem Solving (GPS) has long been a challenging task in the field of artificial intelligence \cite{chou1988introduction,lu2021inter, seo2015solving}, representing a complex form of multimodal mathematical reasoning. Before the rise of LLMs, approaches to GPS were generally divided into two categories: symbolic solvers and neural solvers. 

\textbf{Symbolic solvers} perform precise reasoning by searching for and applying geometric theorems in a structured manner. The InterGPS and Geometry3K datasets \cite{{lu2021inter}} provide high-quality formal language annotations and support theorem-based symbolic reasoning. FormalGeo \cite{zhang2023formalgeo} combines theorem execution with search algorithms, introducing a more diverse set of geometric rules and offering more detailed formal annotations. In addition, Inter-GPS \cite{lu2021inter}, E-GPS \cite{wu2024egps}, GeoDRL \cite{peng2023geodrl} and HyperGNet \cite{zhang2024fgeo} train dedicated neural networks to guide theorem search, while PI-GPS \cite{zhao2025pigps} leverages the capabilities of external LLMs to assist symbolic solvers by predicting theorem sequences and generating structured representations of problem texts. However, all these methods share the inherent limitations of symbolic systems, as they often rely heavily on manually crafted rules \cite{xia2024geox} and formally annotated data, which limits their scalability and adaptability. In contrast, our method also incorporates symbolic systems, but primarily as an enhancement to MLLMs, rather than relying on symbolic reasoning as the core.

\textbf{Neural methods}, on the other hand, adopt end-to-end neural networks to predict program sequences composed of computational operators and elements \cite{lietal2024lans, xiao2024learning}. These sequences are then executed by an external program executor to derive numerical answers. NGS \cite{chen2021geoqa}, DPE-NGS \cite{chen2021geoqa} and Geoformer \cite{chen2022unigeo} design geometry-specific neural modules along with pretraining tasks, and introduce datasets of GeoQA, GeoQA+ and UniGeo. PGPSNet \cite{zhang2023multi} extracts geometric information from diagrams into clauses\cite{hao2022pgdp5k}, enabling better multimodal fusion, and proposes PGPS9K dataset. GeoX \cite{xia2024geox} introduces multi-stage pretraining strategies on ViT and LLM to better align visual features with reasoning, achieving strong performance. While these methods aim to predict program sequences and offer a degree of interpretability, their outputs are still not naturally readable and remain distant from truly natural language solutions \cite{pan2024maths}. In comparison, our approach focuses on enhancing the ability of MLLMs to solve geometry problems directly in natural language, leading to better human readability and greater scalability.

\subsection{MLLMs for Geometry Problem Solving}
Multimodal large language models (MLLMs), such as GPT-4o \cite{hurst2024gpt} and the Qwen VL series \cite{bai2025qwen2}, have demonstrated remarkable capabilities in integrating visual and textual information, opening up new opportunities for GPS task. To adapt these models to domain-specific reasoning tasks, SFT has emerged as the predominant approach. However, the effectiveness of SFT in such tasks is often constrained by the quality of CoT reasoning data \cite{wei2022chain}, which is rarely exist in earlier geometry datasets such as Geometry3K \cite{lu2021inter} and PGPS9K \cite{zhang2023multi}. To overcome data limitations, some approaches explore template-based or LLM-based pipelines to construct large-scale SFT training data. G-LLaVA \cite{gao2023gllava} expands the annotated data in GeoQA \cite{chen2021geoqa} and Geometry3K by leveraging LLMs to rephrase existing annotations. However, it underutilizes other datasets and suffers from limited label diversity. GeomVerse \cite{kazemi2023geomverse} and MAVIS \cite{zhang2024mavis} employ Python engines to generate artificially rendered geometric images. However, these synthetic images still differ significantly from real-world scenarios, and the reasoning paths they provide often lack rich geometric knowledge. Similarly, GeoGPT4V \cite{cai2024geogpt4v} utilizes GPT-4V to generate new problems and corresponding code for synthetic image construction. R-CoT \cite{deng2024rcot} constructs a large number of synthetic images and uses image captions along with an LLM-based Reverse Chain-of-Thought pipeline to generate solution paths. While promising, such LLM-generated data might contain mistakes, and the correctness of the synthesized reasoning steps requires further evaluation. 

In contrast, our GeoGen pipeline offers several advantages: First, it leverages existing symbolic systems and performs precise symbolic reasoning, which enables it to integrate seamlessly with public datasets to fully exploit existing resources, and generate solution paths with guaranteed correctness. Second, GeoGen automatically synthesizes geometric diagrams from formal language expressions, allowing it to represent a wider range of geometric concepts and knowledge.

\section{Methods}

In this section, we present our approach to integrating symbolic systems with MLLMs during both training and inference. As illustrated in Figure  ~\ref{fig:teaser}, our framework consists of two key components: (1) the GeoGen pipeline, which automatically construct CoT reasoning data, and (2) GeoLogic, a bridging module that enables the MLLM to interact with the symbolic system during inference. 

Before detailing our methodology, we first state several key concepts, following previous work \cite{lu2021inter}. A \textit{predicate} denotes a class of geometric relationships, such as \texttt{IsMidpointOfLine}, which indicates that a point is the midpoint of a line segment. A \textit{literal} refers to a instance of a predicate. for example, \texttt{IsMidpointOfLine(P,AB)}, which states that in this geometric diagram, point P is the midpoint of segment AB. 

\subsection{GeoGen Pipeline} 
\label{sec:geogen}

\begin{figure*}[htbp]
  \centering
  \includegraphics[width=\textwidth]{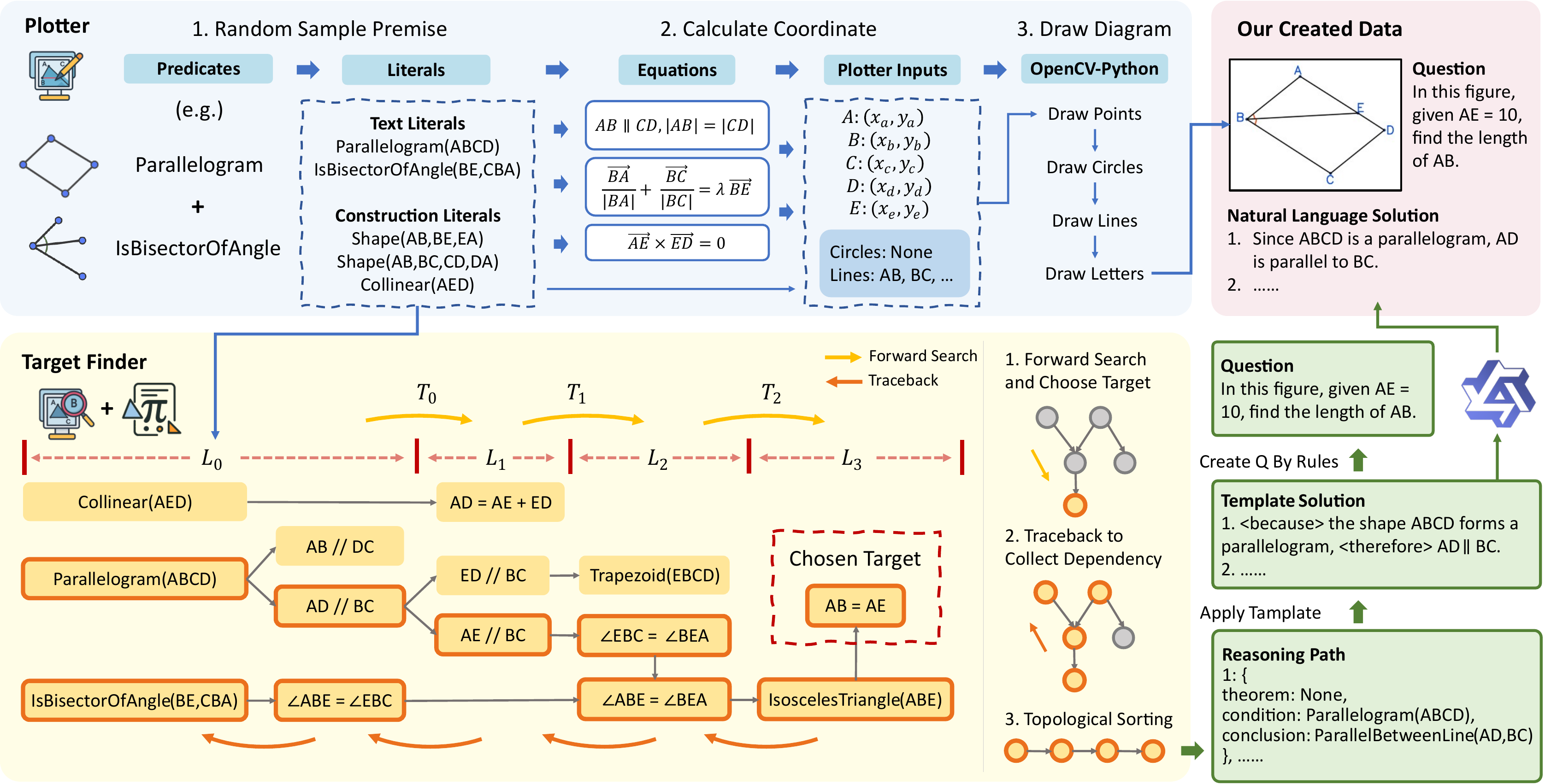}
  \caption{Framework of our proposed GeoGen pipeline.}
  \Description{A block diagram of GeoGen, showing modules such as Plotter, TargetFinder, and how they interact in the pipeline.}
  \label{fig:geogen}
\end{figure*}

GeoGen implements a data synthesis pipeline inspired by AlphaGeometry \cite{AlphaGeometryTrinh2024}, including \textit{random premise sampling}, \textit{symbolic deduction with traceback}, and \textit{synthetic problems}, As shown in Figure \ref{fig:geogen}. In particular, We design the Plotter module to randomly sample geometric conditions and render corresponding diagrams using OpenCV. We also design a Target Finder module that leverages symbolic systems to perform forward search and traceback in order to retrieve solution path. Finally, we apply rule-based generation and external LLMs to convert the structured results into natural language questions and solutions.

\subsubsection{Plotter} 
\label{sec:plotter}

We design a plotter module to perform random premise sampling and automatically generate large-scale geometric diagrams. Specifically, we begin by sampling a combination of geometric predicates, which typically include one entity predicate (e.g. square or isosceles triangle) and one or more relation predicates (e.g. midpoint or median). These predicates are then instantiated into literals by assigning letter labels (e.g., point names like A, B, C) to the corresponding entities. To introduce additional complexity, we further randomly add line segments between sampled points. 

Based on the geometric semantics of each predicate, we construct a group of equations that constraint the coordinates of the relevant points. The coordinates are either directly computed or randomly assigned under the constraints from these equations. Finally, the diagrams are rendered using OpenCV, yielding clean, scalable geometric figures. Since every diagram is generated from a well-defined set of predicates, the resulting diagrams are naturally aligned with geometry predicates and embeds rich geometric knowledge. 

\subsubsection{Target Finder}
\label{sec:targetfinder}

To identify potential goals from geometric images, the Target Finder module performs forward search using a breadth-first strategy. In symbolic system, a theorem becomes applicable when its premises are satisfied by existing literals. Specifically, we begin with an initial set of literals $ L_0 $. By identifying  premises, we obtain a candidate theorem set $ T_0 $. We then apply these theorems to derive new conclusion literals $ L_1 $, which represent the first layer of inferred conditions. Based on $ L_1 $ we can find new theorems $ T_1 $ applicable. Repeating this process gradually uncovers deeper layers of geometric relationships until all possible conditions are discovered. Additionally, we record the premises that led to each literal during forward search. This process can constructs a directed graph, where nodes represents literals and edges represent their logical dependencies.

After the constructoin of directed graph, we select a node as the target and trace backward through the directed graph to identify all its dependencies. Typically, nodes located deeper in the graph are involved in longer reasoning chains and more complex targets. After filtering out overly complex nodes, we randomly select one or more suitable nodes as targets. The corresponding subgraph is then extracted, and topological sorting is applied to linearize the reasoning subgraph, resulting in a sequence of (condition, theorem, conclusion) triples that reflect the logical structure of geometric reasoning. Finally, we apply template-based rewriting and leverage an external LLM to refine the sequence into clear and diverse chain-of-thought (CoT) explanations.

\subsection{Dataset Construction via GeoGen}

To construct high-quality training data for geometry problem solving, we leverage our proposed GeoGen pipeline to automatically generate CoT data for both existing and synthetic geometry diagrams. Many existing geometry datasets lack natural language CoT explanations, making them suboptimal for training MLLMs. We primarily focus on expanding two widely used datasets: Geometry3K \cite{lu2021inter} and PGPS9K \cite{zhang2023multi}. Compared to other datasets like GeoQA+ \cite{geoqa_plus}, which already has a high-quality extended dataset (Geo170K) \cite{gao2023gllava}, or proving part of UniGeo \cite{chen2022unigeo}, which consists of geometry proofs without formal symbolic annotations or structured reasoning steps, Geometry3K and PGPS9K offer a more suitable foundation for augmentation using our GeoGen pipeline.


First, GeoGen explores multiple symbolic reasoning paths for each image. For each derived path, we design two types of questions: one includes a description of the geometric figure, while the other provides only the target to be solved. This dual design enables both vision-guided reasoning and text-only inference, enriching the diversity of training samples. In addition, we introduce proof-style questions to further increase the complexity and variety of reasoning. As a result, we generate over 45,526 QA pairs from 4,849 original images in the Geometry3K and PGPS9K datasets.

To further scale up the training data, we use GeoGen’s Plotter module to synthesize geometry diagrams, and create CoT data using the same method. As discussed above in Section ~\ref{sec:plotter}, Plotter starts by randomly sampling a set of predicates. We varied both the number and types of predicates to generate diverse diagram structures. Specifically, we explored 3,159 unique predicate combinations, producing 129,230 diagrams in total. To maintain visual clarity and avoid overly complex images, we limited the number of predicates per diagram. Additionally, we observed that many diagrams tended to result in similar theorem usage patterns during generation. To promote knowledge diversity, we capped the number of Q\&A pairs per unique theorem sequence to 400 and removed redundant questions, resulting in the final 62,868 selected samples.

We further analyze the distribution of reasoning steps in our GeoExpand and GeoSynth, which reflects the depth of symbolic inference required, as shown in Figure ~\ref{fig:teaser}. Although longer reasoning chains may not always correspond to more difficult problems, this distribution gives a rough estimate of the dataset's complexity. 

\subsection{Model Training}

We train our model on multiple MLLMs that combine a Vision Transformer (ViT) \cite{dosovitskiy2020image} with a large language model. We adopt a single-stage training paradigm, where the vision encoder is kept frozen and only the language module is updated. This simple training setup is intended to highlight the effectiveness of our data, independent of complex fine-tuning strategies.

We adopt a standard auto-regressive language modeling objective for SFT in the multimodal setting. Given an image input $ I $ and a corresponding target output sequence $ y = (y_1, y_2, \ldots, y_T) $, the training loss is defined as:

\begin{equation}
  \mathcal{L}_{\text{SFT}} = - \sum_{t=1}^{T} \log P_\theta\left( y_t \mid y_{<t}, \mathbf{x}_{\text{text}}, \mathbf{x}_{\text{image}} \right)
\end{equation}

Here, $ y_t $ denotes the target token at time step $ t $, $ y_{<t} $ represents the sequence of previously generated tokens, and $ \mathbf{x}_{\text{text}} $ represents input prompt. The model is trained to maximize the likelihood of the correct next token $ y_t $ conditioned on both the image $ \mathbf{x}_{\text{image}} $ and the previous text. 

\subsection{Reasoning with GeoLogic}

To better integrate symbolic reasoning with the capabilities of MLLMs, we introduce GeoLogic, a LLM trained to bridge the gap between the natural language reasoning steps predicted by MLLMs and the formal language required by symbolic systems. GeoLogic assists MLLMs in producing logic-grounded reasoning.

As discussed in Section ~\ref{sec:targetfinder}, we use an external LLM to convert structured reasoning paths into natural language, resulting in paired data of formal and natural language explanations. We then decompose these explanations into step-by-step reasoning units, and train GeoLogic using SFT with a standard autoregressive SFT loss, where the input is a natural language reasoning step and the output is its corresponding formal logic representation.

During inference, GeoLogic acts as a translator: it converts each natural language reasoning step predicted by the MLLM into a formal triple of condition, theorem and conclusion, which is then verified by a symbolic reasoning system. We explore two verification modes. Strict mode checks whether the condition, theorem, and conclusion all match valid symbolic reasoning outputs. However, symbolic theorem application requires strictly correct inputs—including naming and ordering of geometric entities—which imposes significant engineering challenges. We also implement fast mode, which only checks whether the necessary visual elements for a conclusion exist, rather than verifying the correctness of the conclusion itself. For example, given a model prediction like \texttt{IsMidsegmentOfTriangle(DE,ABC)}, we only check whether segment DE and triangle ABC are present in the diagram. Although this mode cannot confirm logical correctness, it helps ensure the model's predictions are grounded in visual evidence and prevents hallucination of nonexistent geometric objects.

We train the MLLM and the GeoLogic model using the data generated by GeoGen. During inference, we adopt a step-level tree search algorithm that integrates symbolic verification to guide the MLLM's reasoning process. At each iteration, the MLLM generates $ K $ candidate reasoning steps in natural language. These candidates are then translated into formal language using the GeoLogic. Each formal step is verified by the symbolic system, which returns a boolean result indicating whether the step is valid. Among the valid steps, one is randomly selected and appended to the input prompt as part of the reasoning history. The process repeats, progressively expanding the reasoning chain, until the model produces no further output. Our step-level tree search algorithm is as follows:

\begin{algorithm}[ht]
  \caption{Step-Level Tree Search with Symbolic Verification}
  \begin{algorithmic}[1]
  \State \textbf{Input:} Initial prompt $P$, MLLM model $M$, GeoLogic translator $G$, symbolic system $S$, number of candidates $K$
  \State \textbf{Initialize:} Reasoning history $H \gets \emptyset$
  \While{True}
      \State Candidate steps $C \gets M.\text{GenerateSteps}(P, K)$
      \State Formal language $F \gets [G.\text{Translate}(c) \ \textbf{for} \ c \in C]$
      \State Verification results $V \gets [S.\text{Verify}(f) \ \textbf{for} \ f \in F]$
      \State $C_{\text{valid}} \gets \{c_i \in C \mid V[i] = \text{True}\}$
      \If{$C_{\text{valid}} = \emptyset$}
          \State \textbf{break}
      \EndIf
      \State Randomly select $c^* \in C_{\text{valid}}$
      \State Append $c^*$ to history: $H \gets H \cup \{c^*\}$
      \State Update prompt: $P \gets P \oplus c^*$
  \EndWhile
  \State \textbf{Return} reasoning history $H$
  \end{algorithmic}
  \end{algorithm}

\section{Experiments}

\begin{table*}[ht]
  \resizebox{\textwidth}{!}{%
  \begin{tabular}{c|cc|cccc|cccc}
  \hline
  \multirow{2}{*}{Model}          & \multirow{2}{*}{Geo170K} & \multirow{2}{*}{G-Exp \& G-Syn} & \multicolumn{4}{c|}{Completion Acc (\%)}           & \multicolumn{4}{c}{Choice Acc (\%)}                \\
                                  &                          &                              & GeoQA & Geometry3K & PGPS9K & \textbf{Avg.}            & GeoQA & Geometry3K & PGPS9K & \textbf{Avg.}           \\ \hline
  \multirow{3}{*}{InternVL2.5-2B} &                          &                              & 35.4 & 39.1      & 34.6  & 36.0            & 55.6 & 52.6      & 51.0  & 52.9           \\
                                  & \ding{51}                 &                              & 59.2 & 18.1      & 16.1  & 30.4            & 69.6 & 42.8      & 43.1  & 51.5           \\
                                  & \ding{51}                 & \ding{51}                    & 54.6 & 42.1      & 36.8  & \textbf{43.9}   & 68.7 & 51.1      & 44.6  & \textbf{54.0}  \\ \hline
  \multirow{3}{*}{Qwen2.5-VL-3B}  &                          &                              & 28.7 & 31.8      & 24.9  & 27.9            & 56.8 & 41.4      & 31.6  & 42.2           \\
                                  & \ding{51}                &                              & 60.2 & 21.3      & 21.3  & 33.8            & 72.8 & 46.1      & 43.6  & 53.6           \\
                                  & \ding{51}                & \ding{51}                      & 61.4 & 38.6      & 36.7  & \textbf{45.1}   & 75.6 & 51.1      & 44.7  & \textbf{56.2}   \\ \hline
  \multirow{3}{*}{LLaVA-v1.6-7B}  &                          &                              & 13.8 & 12.7      & 9.4   & 11.6            & 17.9 & 19.8      & 20.1  & 19.3           \\
                                  & \ding{51}              &                              & 51.3 & 4.5       & 11.6  & 22.5            & 61.1 & 32.5      & 28.7  & 40.0           \\
                                  & \ding{51}                 & \ding{51}                    & 50.7 & 34.8     & 37.4  & \textbf{41.0}  & 60.0 & 42.9      & 43.1  & \textbf{48.5}    \\ \hline
  \multirow{3}{*}{Qwen2.5-VL-7B}  &                          &                              & 64.2 & 44.8      & 43.0  & 50.0             & 64.3 & 56.4      & 48.3  & 55.5           \\
                                  & \ding{51}                 &                              & 65.8 & 27.5      & 23.0  & 37.8            & 76.5 & 52.4      & 47.2  & 57.9           \\
                                  & \ding{51}                & \ding{51}                    & 64.9 & 46.3      & 43.9  & \textbf{51.2}   & 77.6 & 58.4      & 54.3  & \textbf{62.9}   \\ \hline
  \end{tabular}%
  }
  \caption{ Ablation study on different training configurations across MLLMs under both Completion and Choice settings. G-Exp and G-Syn denote the GeoExpand and GeoSynth datasets, respectively.}
  \label{tab:ablation_model}
  \end{table*}

\subsection{Setup}

\subsubsection{Dataset}

As previously discussed, we constructed GeoExpand by extending the Geometry3K and PGPS9K datasets, resulting in 45K samples, and created GeoSynth from scratch with 62K samples as part of our training set. In addition, to prevent performance degradation on other benchmark datasets, we included the instruction-tuning part of Geo170K during training, which consists of over 117K Q\&A instruction pairs. This dataset is primarily derived from GeoQA+, which is complementary to our work and enhances the model’s robustness across diverse types of geometric problems. In total, our training set comprises over 224K samples.

\subsubsection{Implementation Details}

For data synthesis in the GeoGen pipeline, we employ the API of Qwen2.5-32B-Instruct to transcribe the generated template solution using a one-shot prompting strategy. This is a relatively straightforward task for a powerful LLM. To verify transcription quality, we randomly sampled 100 generated natural language solutions and manually checked them, finding no factual errors. Our symbolic system is based on FormalGeo, which defines 88 geometric predicates and 196 theorems. We further restructured the equation solver in FormalGeo to produce more detailed and interpretable solution steps. In addition, we manually expanded the original theorem set to include 210 theorems, further enriching the expressive power of the system.

For model training, all experiments are conducted based on PyTorch 2.5.1 with CUDA 11.8, running on a single node with 8 NVIDIA A40 GPUs (each with 48GB memory). We use a consistent learning rate of 1e-5 with gradient accumulation to maintain a global batch size of 16. A cosine learning rate schedule is applied, including a warm-up phase covering 5\% of the total training steps. Our main experiments are trained for 3 epochs. We employ the AdamW optimizer with $ \beta_1 = 0.9 $ and $ \beta_2 = 0.95 $.

\subsubsection{Evaluation Metric}

We evaluate model performance on the test sets of GeoQA, Geometry3K, and PGPS9K, which contain 754, 601, and 1,000 samples respectively. We report the overall accuracy across these 2,355 samples, and refer to this combined evaluation set as GeoTest. Additionally, we evaluate our model on a subset of geometry problems from the MathVista mini set \cite{lu2023mathvista}, which consists of 208 samples collected from Geometry3K, GeoQA+, UniGeo, and GEOS \cite{seo2015solving}. Due to the relatively small sample size, the evaluation results on this set are subject to high variance.

We adopt Top-1 accuracy as our primary evaluation metric, rather than Top-3 or Top-10 accuracy that symbolic or neural baselines \cite{lu2021inter,chen2021geoqa} often report. Top-K evaluation strategy considers a prediction correct if \textit{any} of the Top-K answers match the ground truth. However, such a metric is less suitable for MLLMs, as these models can often reach the correct answer after multiple attempts, potentially leading to an overestimation of their actual performance.  In contrast, Top-1 accuracy, which requires the model to produce the correct answer in a single response, provides a more faithful reflection of its reasoning ability.

Following prior work \cite{gao2023gllava, lietal2024lans}, we evaluate models under two settings: Completion, which requires the model to directly generate the final answer; and Choice, which involves selecting the correct answer option from a given set. For completion, we adopt a prompting strategy with Qwen2.5-7B \cite{yang2024qwen2} as an automatic answer verifier. Given the reference (label) answer and the final three sentences from the MLLM’s response, the checker determines whether the model's predicted numerical answer is correct. For choice, we apply regular expression matching to extract the selected option (e.g., A, B, C, D) from the model's output. A prediction is marked incorrect if parsing fails or the extracted choice differs from the ground truth.

\subsection{Cross-Model Effectiveness}


We conduct a comprehensive ablation study across multiple MLLM families and model sizes, as presented in Table~\ref{tab:ablation_model}. Compared to training solely on Geo170K, incorporating our data consistently boosts performance across different architectures and parameter scales. This highlights the robustness and wide applicability of our data augmentation strategy. Notably, some models even show a drop in performance when fine-tuned only on Geo170K, suggesting that relying on a single-source dataset may hinder generalization—particularly across diverse benchmarks. In contrast, our data introduces a richer variety of geometric structures and reasoning patterns, effectively complementing Geo170K and leading to more generalized improvements across evaluation settings.

\begin{table*}[htbp]
  \centering
  \resizebox{\textwidth}{!}{%
  \begin{tabular}{l|cccc|cccc}
  \hline
  \multirow{2}{*}{\textbf{Methods}} & \multicolumn{4}{c|}{Completion Acc (\%)} & \multicolumn{4}{c}{Choice Acc (\%)} \\
                                    & GeoQA & Geometry3K & PGPS9K & MathVista-min-GPS           & GeoQA & Geometry3K & PGPS9K & MathVista-mini-GPS \\
  \hline
  \multicolumn{9}{c}{\textit{Symbolic-based Methods}} \\
  \hline
  InterGPS* \cite{lu2021inter}                       & –     & 44.6       & –      & –                        & –     & 56.9       & –      & – \\
  Pi-GPS*  \cite{zhao2025pigps}                        & –     & 70.6       & 61.4   & –                        & –     & 77.8       & 69.8   & – \\
  \hline
  \multicolumn{9}{c}{\textit{Neural Methods}} \\
  \hline
  NGS* \cite{chen2021geoqa}                            & 60.0  & 35.3       & 34.1   & –                        & –     & 58.8       & 46.1   & – \\
  Geoformer* \cite{chen2022unigeo}                      & 62.5  & 36.8       & 35.6   & –                        & –     & 59.3       & 47.3   & – \\
  PGPSNet-v2-S* \cite{zhang2024fuse}                    & –     & 65.2       & 60.3   & –                        & –     & 76.4       & 69.2   & – \\
  GeoX* \cite{xia2024geox}                           & 54.9  & 58.6       & 52.7   & –                        & –     & 72.5       & 63.3   & – \\
  \hline
  \multicolumn{9}{c}{\textit{Open Source General MLLMs}} \\
  \hline
  Qwen2.5-VL-3B \cite{bai2025qwen2}                   & 28.7 & 31.8     & 24.9   & 35.6                  & 56.8 & 41.4      & 31.6  & 42.2 \\
  LLaVA-NeXT-7B \cite{liu2024llavanext}                   & 13.8 & 12.7     & 9.4   & 10.6                  & 17.9 & 19.8      & 20.1  & 32.2 \\
  Qwen2-VL-7B \cite{qwen2}                     & 43.9 & 32.5      & 24.3  & 27.4                    & 46.2 & 42.3      & 35.0  & 49.0 \\
  Qwen2.5-VL-7B \cite{bai2025qwen2}                   & \underline{64.2} & \underline{44.8}      & \underline{43.0}  & \underline{54.3}                    & 64.3 & \underline{56.4}      & \underline{48.3}  & 56.7 \\
  InternVL2.5-8B \cite{chen2024internvl}                 & 50.5 & 41.4      & 26.0  & 47.1                    & 58.1 & 52.6      & 37.4  & 54.3 \\
  \hline
  \multicolumn{9}{c}{\textit{Geometry MLLM Solvers}} \\
  \hline
  G-LLaVA-13B* \cite{gao2023gllava}                    & –     & –          & –      & –                        & 67.0  & –          & –      & 56.7 \\
  MAVIS-7B* \cite{zhang2024mavis}                       & –     & –          & –      & –                        & 68.3  & –          & –      & 64.1 \\
  EAGLE-7B* \cite{li2024eagle}                       & –     & –          & –      & –                        & 67.1  & –          & –      & 54.3 \\
  R-CoT-8B* \cite{deng2024rcot}                       & –     & –          & –      & –                        & \underline{75.9}  & –          & –      & \underline{73.1} \\
  \textbf{GeoGen-SFT-3B}             & 61.4 & 38.6      & 36.7  & 49.5                   & 75.6 & 51.1     & 44.7  & 64.5 \\
  \textbf{GeoGen-SFT-7B}             & \textbf{64.6} & \textbf{46.3} & \textbf{43.9} & \textbf{63.9} & \textbf{78.0} & \textbf{58.4} & \textbf{54.3} & \textbf{74.0} \\
  \hline
  \end{tabular}
  } 
  \caption{Our experimental results along with the current state-of-the-art results from different categories of methods under both Completion and Choice settings. Results marked with * are reported from original papers. \textbf{Bold} numbers indicate the best results, and \underline{underlined} numbers denote the second-best.}
  \label{tab:sota}
\end{table*}

\subsection{Data Composition Analysis}

\begin{table}[t]
  \centering
  \begin{tabular}{c|c|c|c|c|c}
  \hline
  & Geo170K & G-Exp(1) & G-Exp(2) & G-Syn & Training Size \\
  \hline
  T0 & \ding{51} &         &         &        & 117k \\
  T1 & \ding{51} & \ding{51} &         &        & 141k \\
  T2 & \ding{51} & \ding{51} & \ding{51} &      & 162k \\
  T3 & \ding{51} & \ding{51} & \ding{51} & \ding{51} & 224k \\
  \hline
  \end{tabular}
  \caption{Composition of training datasets used in the Figure ~\ref{fig:geo_ablation_2}. G-Exp(1) and G-Exp(2) refer to the GeoExpand subsets derived from Geometry3K and PGPS9K, respectively. G-Syn denotes the GeoSynth dataset.}
  \label{tab:geo_ablation_2}
  \end{table}

\begin{figure}[t]
  \centering
  \includegraphics[width=\columnwidth]{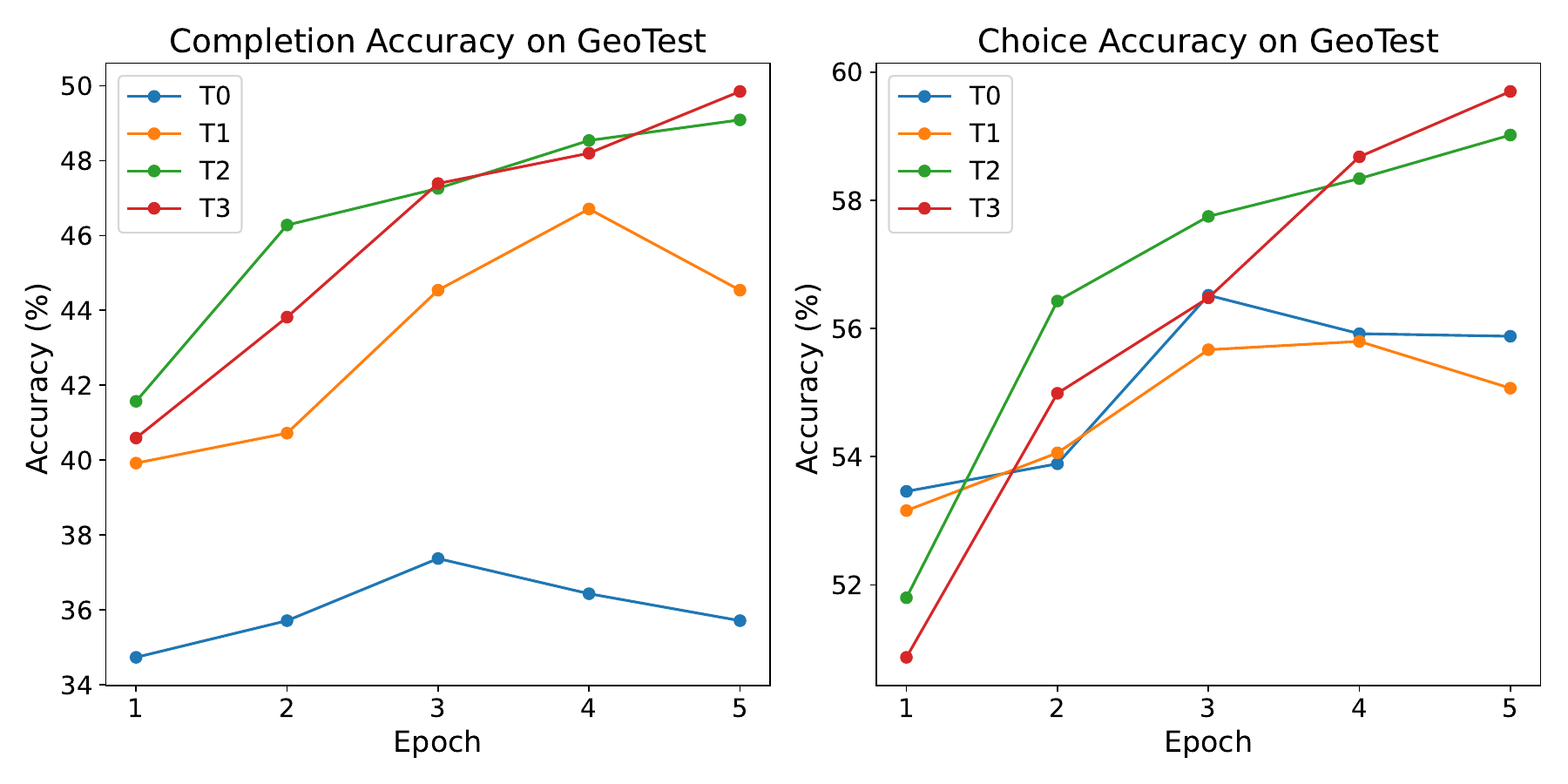}
  \caption{We conduct ablation study to examine the impact of different training data compositions. This figure shows the performance on GeoTest across epochs for each configuration. T0-T3 settings are defined in Table ~\ref{tab:geo_ablation_2}.}
  \Description{We conduct ablation study to examine the impact of different training data compositions. The line chart shows the performance on GeoTest across epochs for each configuration.}
  \label{fig:geo_ablation_2}
  
\end{figure}

We conduct a detailed ablation study on Qwen2.5-VL-3B to investigate the impact of different data components in our training corpus with evaluation ion GeoTest. As shown in Table~\ref{tab:geo_ablation_2}, we define four training configurations (T0–T3), gradually incorporating subsets from our proposed datasets: GeoExpand (G-Exp) and GeoSynth (G-Syn). Specifically, T0 uses only the original Geo170K instruction-tuning dataset, while T3 combines all available components, including both parts of GeoExpand and GeoSynth. To account for the varying sizes of training sets and their effects on convergence speed, we evaluate model performance at each epoch during training. The accuracy curves for both Completion and Choice settings are illustrated in Figure~\ref{fig:geo_ablation_2}.

From the plots, we observe several key trends. First, the best performance of each configuration improves consistently with the increase in data volume, highlighting the benefits of data augmentation. Both Completion and Choice accuracies steadily rise from T0 to T3, demonstrating the effectiveness of our GeoExpand and GeoSynth datasets. Notably, GeoExpand contributes substantial early gains—comparing T0, T1, and T2, we observe clear improvements as we incrementally add data from Geometry3K and PGPS9K, particularly in the initial training epochs. GeoSynth further boosts long-term performance. Although the early-stage gains from T2 to T3 are less dramatic, the final accuracy after five epochs shows that GeoSynth enables better generalization, ultimately leading to the highest overall performance.

\subsection{Compared with State-of-the-art Methods}

Table~\ref{tab:sota} presents a comprehensive comparison of our method against existing state-of-the-art models across four geometry benchmarks, under both completion and choice settings. Due to the fundamental differences in methodology, we primarily compare MLLM-based approaches. Results of symbolic-based methods and neural solvers are also reported for reference.

We fine-tune both Qwen2.5-VL-3B and Qwen2.5-VL-7B on data of Geo170K and our GeoExpand, GeoSynth, denoted by the prefix GeoGen-SFT in Table~\ref{tab:sota}. Among MLLM-based methods, our 7B model achieves the best performance across all geometry datasets, reaching 63.9\% completion accuracy and 74.0\% choice accuracy on the MathVista-mini-GPS dataset. In particular, it attains 78.0\% accuracy under the choice setting on the widely-used GeoQA benchmark, outperforming all existing MLLM-based geometry solvers. It consistently surpasses both general-purpose MLLMs and geometry-focused MLLM solvers, demonstrating the effectiveness of domain-specific data and symbolic supervision. Furthermore, our method shows comparable or even better performance than some earlier symbolic or neural methods, such as InterGPS and NGS. However, the latest symbolic and neural solvers still significantly outperform current MLLM-based approaches, highlighting that there remains substantial room for improvement in the geometric reasoning capabilities of MLLMs.

\subsection{Tree Search Integrating Symbolic System}

\begin{figure}[t]
  \centering
  \includegraphics[width=0.8\columnwidth]{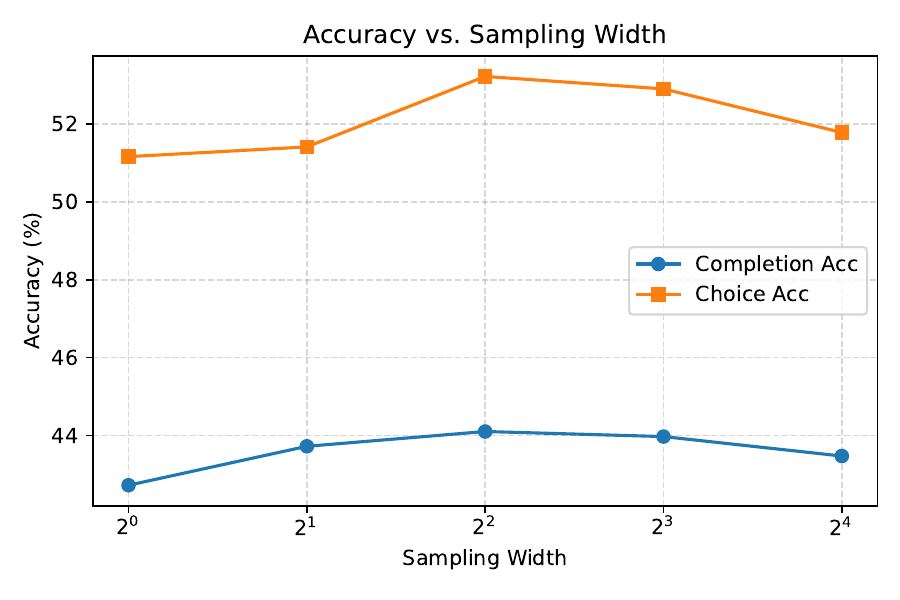}
  \caption{Accuracy trends as we vary the search width during symbolic reasoning in the inference stage. We adopt the same evaluation metrics as before.}
  \Description{Experimental results on varying the search width.}
  \label{fig:search_result}
  
\end{figure}

We first trained the GeoLogic model based on Qwen2.5-3B, achieving a prediction accuracy of 94.16\% on our custom test set. A prediction by GeoLogic is considered correct only if it exactly matches the target string. Building on this, we incorporated a symbolic system during the inference stage to enable tree-based search and automatic verification of reasoning steps. We varied the search width to evaluate its impact and conducted experiments using the Qwen2.5-VL-7B model trained on both GeoExpand and GeoSynth. The evaluation was conducted on a combined test set of Geometry3K and PGPS9K. The results are shown in Figure~\ref{fig:search_result}.

As observed, model performance initially improves as the search width increases, suggesting that symbolic verification helps filter out incorrect reasoning paths and enhance prediction quality. However, further increasing the width leads to a decline in performance. One possible explanation is that our verification process only checks for necessary conditions of a conclusion. Although a model’s prediction may satisfy these conditions, it does not guarantee the sufficiency or overall correctness of the conclusion itself. A wider search yields more candidate paths that pass verification, and randomly selecting among them may approximate the effect of random sampling. The comparable results between width 1 and width 16 further support this interpretation. 

However, the reasoning paths obtained through symbolic verification are more reliable, as they consistently remain aligned with the geometric information in the image, which will be discussed in Section ~\ref{sec:case_study}. Overall, integrating symbolic verification into the inference stage represents only an initial step. Enabling MLLMs to work more effectively with symbolic systems remains a significant challenge for future research.

\section{Case Study}

\label{sec:case_study}
\begin{figure}[t]
  \centering
  \includegraphics[width=\columnwidth]{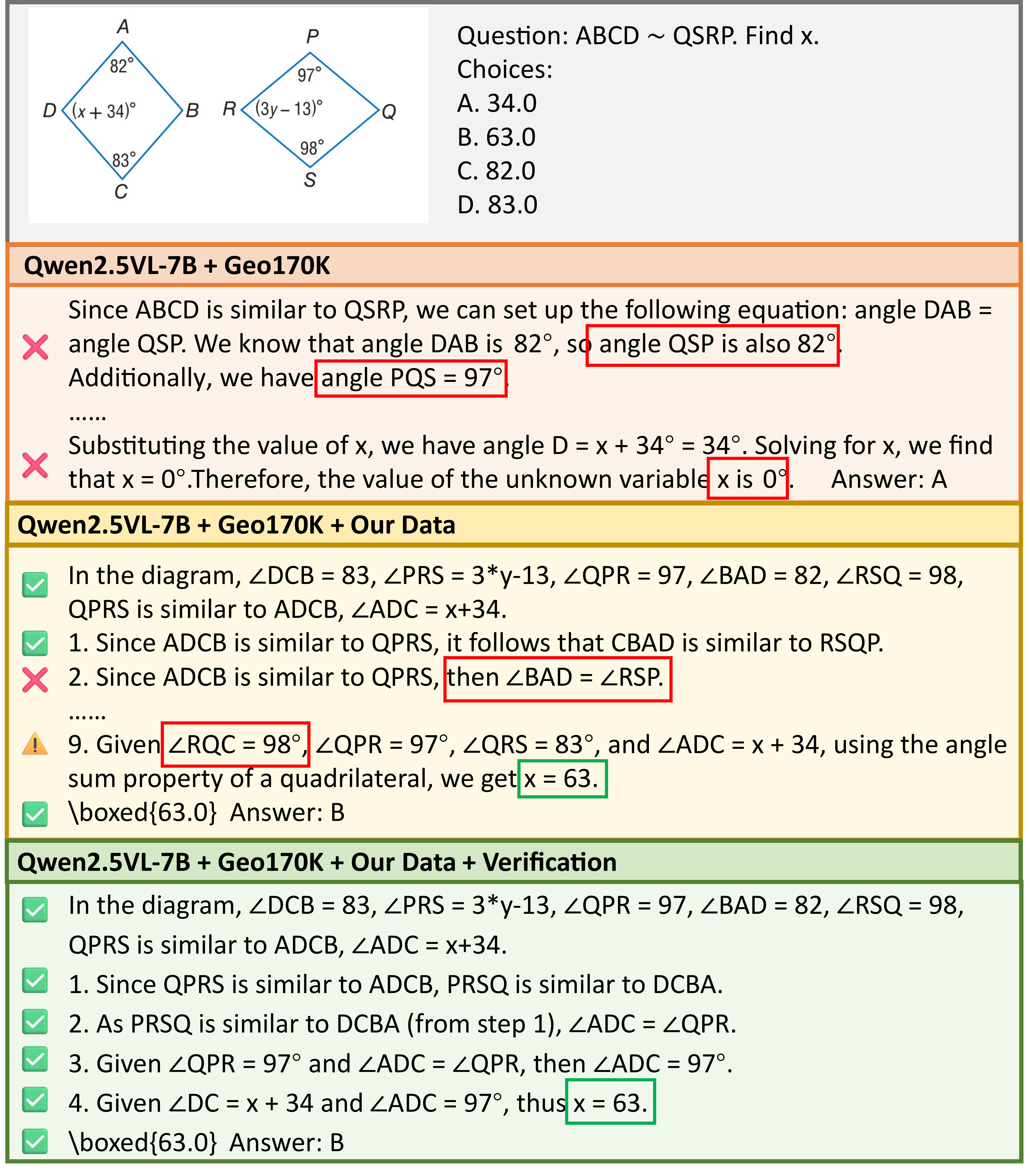}
  \caption{A typical case with model predictions improving as our methods are progressively integrated.}
  \Description{A typical case with model predictions improving as our methods are progressively integrated.}
  \label{fig:case_study}
  
\end{figure}

As illustrated in Figure~\ref{fig:case_study}, we present a representative geometry problem to demonstrate how model predictions evolve as our methods are progressively integrated. When trained solely on Geo170K, the model struggles with reasoning errors and frequently hallucinations that are misaligned with the diagram. After incorporating our synthesized data from GeoGen, the model shows improved step-by-step reasoning and successfully predicts the correct answer. However, the reasoning path still contains inconsistencies, revealing that the solution is not fully reliable. These issues are difficult to capture through answer-only evaluation metrics. By further integrating GeoLogic and the symbolic system during inference, the model generates more accurate and coherent reasoning steps. This symbolic guidance mitigates hallucinations, enhances logical consistency, improves alignment with visual content, and significantly boosts the interpretability of the reasoning process.

\section{Conclusion}

In this work, we propose GeoGen, a novel pipeline for geometry problem solving that integrates symbolic reasoning with multimodal large language models. GeoGen enables the large-scale synthesis of step-wise reasoning data grounded in geometric diagrams. We construct two extensive multi-step reasoning datasets and train both an MLLM and a GeoLogic model, which bridges natural language reasoning with formal symbolic logic. Experimental results across multiple benchmarks demonstrate that our approach consistently improves reasoning accuracy, mitigates hallucinations, and enhances the interpretability of the reasoning process. In future work, we plan to further enhance the verification capabilities of the symbolic system, develop more effective interaction mechanisms between MLLMs and symbolic tools, and explore their application in reinforcement learning scenarios.

\bibliographystyle{ACM-Reference-Format}
\bibliography{sample-base}










\end{document}